\documentclass[11pt]{article}

\usepackage[]{acl}
\usepackage{times}
\usepackage{latexsym}
\usepackage{subcaption}
\usepackage{adjustbox}
\usepackage{booktabs}
\usepackage[T1]{fontenc}
\usepackage[utf8]{inputenc}

\usepackage{microtype}

\usepackage{inconsolata}

\usepackage{graphicx}

\usepackage{xcolor}
\usepackage{colortbl}
\usepackage{booktabs}
\usepackage{pgfplotstable}
\pgfplotsset{compat=1.18}

\title{The LSCD Benchmark: a Testbed for Diachronic Word Meaning Tasks}

\author{Dominik Schlechtweg\textsuperscript{1}, Sachin Yadav\textsuperscript{1}, Jonas Kuhn\textsuperscript{1}, Nikolay Arefyev\textsuperscript{2}\\
University of Stuttgart\textsuperscript{1}, University of Oslo\textsuperscript{2}\\
\texttt{first.last@\{ims.uni-stuttgart.de\}}
}
  
\begin{document}
\maketitle
\begin{abstract}
Lexical Semantic Change Detection (LSCD) is a complex, lemma-level task, which is usually operationalized based on two subsequently applied usage-level tasks: First, Word-in-Context (WiC) labels are derived for pairs of usages. Then, these labels are represented in a graph on which Word Sense Induction (WSI) is applied to derive sense clusters. Finally, LSCD labels are derived by comparing sense clusters over time. This \textbf{modularity} is reflected in most LSCD datasets and models. It also leads to a large \textbf{heterogeneity} in modeling options and task definitions, which is exacerbated by a variety of dataset versions, preprocessing options and evaluation metrics. This heterogeneity makes it difficult to evaluate models under comparable conditions, to choose optimal model combinations or to reproduce results. Hence, we provide a benchmark repository standardizing LSCD evaluation. Through transparent implementation results become easily reproducible and by standardization different components can be freely combined. The repository reflects the task's modularity by allowing model evaluation for WiC, WSI and LSCD. This allows for careful evaluation of increasingly complex model components providing new ways of model optimization.
\end{abstract}

\section{Introduction}
\label{sec:intro}

Lexical Semantic Change Detection (LSCD) is a field of NLP that studies methods automating the analysis of changes in word meanings over time. In recent years, this field has seen much development in terms of models, datasets and tasks \citep{surveysemanticchange,periti2024lexical}, motivated by applications in historical linguistics or the digital humanities for theory testing \citep{Hamilton:2016,karjus2025machine}, or in lexicography for practical dictionary maintenance \citep{Skoldberg2024revealing}. 
LSCD is a complex, lemma-level task, which is usually operationalized based on two subsequently applied usage-level tasks: First, Word-in-Context (WiC) labels are derived for pairs of usages. Then, these labels are represented in a graph on which Word Sense Induction (WSI) is applied to derive sense clusters. Finally, LSCD labels are derived by comparing sense clusters over time. This \textbf{modularity} is reflected in most LSCD datasets and models. It also leads to a large \textbf{heterogeneity} in modeling options and task definitions, which is exacerbated by a variety of dataset versions, preprocessing options and evaluation metrics. This heterogeneity makes it difficult to evaluate models under comparable conditions, to choose optimal model combinations or to reproduce results. 

In order to handle this heterogeneity, we think that a shared testbed with a common evaluation setup is needed. Hence, we present a benchmark repository implementing evaluation procedures for models on most available LSCD datasets.\footnote{Find the code at \url{https://github.com/Garrafao/LSCDBenchmark}.} 
The benchmark exploits the modularity of the meta task LSCD by allowing for evaluation of the subtasks WiC and WSI on the same datasets. It can be assumed that performance on the subtasks directly determines performance on the meta task. We aim to stimulate transfer between the fields of WiC, WSI and LSCD by providing a repository allowing for evaluation on all these tasks with shared model components.

\begin{figure*}[t]
  \centering
  \includegraphics[width=0.99\textwidth]{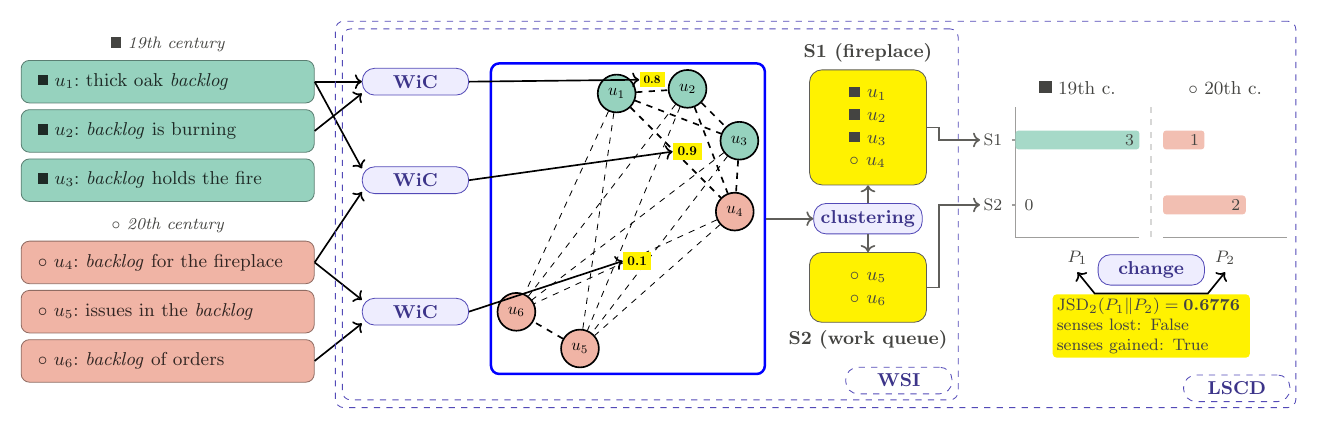}
  \caption{The straightforward approach to LSCD mimicking the annotation process of the popular LSCD datasets.}
  \label{fig:runningex}
\vspace{-15pt}
\end{figure*}

\section{Related Work}
\label{sec:related}

A number of recently created LSCD datasets apply WiC and WSI in the annotation process (cf. Section \ref{sec:tasks}) and thus allow for evaluation of WiC and WSI along with LSCD models \citep[i.a.][]{Kurtyigit2021discovery,Schlechtweg2021dwug,kutuzov2022nordiachange,Zamora2022lscd,Chen2023chiwug}.\footnote{The bulk of these datasets is listed at: \url{https://www.ims.uni-stuttgart.de/data/wugs}.} 
There are also a number of datasets omitting WSI, but allowing for WiC and LSCD evaluation \citep[i.a.][]{Schlechtwegetal18,rodina2020rusemshift,rushifteval2021}, or datasets omitting the WiC allowing for WSI and LSCD evaluation \citep[i.a.][]{diacrita_evalita2020,DBLP:conf/coling/CookLMB14}, or datasets allowing purely for WiC evaluation \citep[][]{loureiro-etal-2022-tempowic}. 

So far, there is no comprehensive LSCD benchmark, implementing SOTA models on (human-annotated) high-quality evaluation data from multiple languages and multiple time periods. The leaderboards of several standalone shared tasks can be seen as small-scale benchmarks without common model implementation \citep[][]{AhmadEtal2020,diacrita_evalita2020,rushifteval2021,Zamora2022lscd,fedorova2024axolotl} with the SemEval task being the most diverse with four languages \citep[][]{schlechtweg-etal-2020-semeval}. \citet[][]{Schlechtwegetal19} provide a comprehensive repository of type-based modeling approaches to LSCD with evaluation pipelines on multiple datasets. 
However, type-based models have more recently been outperformed by token-based contextualized embedding approaches \citep[][]{rushifteval2021,Zamora2022lscd} and especially WiC-fine-tuned models \citep[][]{Cassotti2023Xllexeme,deepmistake-lcsdiscovery}. LLMs have shown competitive performance on some datasets \citep{zamorareina2025lscd} while being less efficient limiting their applicability in large data scenarios. \citet{Periti2024systematic} perform a systematic comparison, but do not provide a flexible model implementation and do not evaluate on some of the most recent SOTA models or datasets.

\section{Tasks}
\label{sec:tasks}

LSCD can be seen as the combination of (at least) three lexical semantic tasks \citep[][]{Schlechtweg2023measurement}: (i) measurement of semantic proximity between word usages, (ii) clustering of the usages based on their semantic proximity, and (iii) estimation of semantic change labels from the obtained clusterings. Task (i) and (ii) corresponds to the lexicographic process of deriving word senses \citep{Kilgarriff2006}, while task (iii) measures LSC based on the derived word senses. The tasks need to be solved sequentially, in the order given above, as each is dependent on the output of the previous task, e.g., word usages can only be clustered once their semantic proximity has been estimated. 

The three tasks are reflected in the human \citep[e.g.][]{schlechtweg-etal-2020-semeval,Schlechtweg2021dwug,kutuzov2022nordiachange} as well as the computational process \citep[e.g.][]{giulianelli-etal-2020-analysing,Montariol2021scalable,Laicher2021explaining,deepmistake-lcsdiscovery} of measuring lexical semantic change.\footnote{However, it is not always obvious as annotation and modeling procedures often try to simplify or skip steps of this process.} 
The first task is known as a standalone task under the name of "Word-in-Context" \citep[WiC, ][]{pilehvar2019wic} while the second task is known as the task of "Word Sense Induction" \citep[WSI, ][]{Schutze1998}. A number of recently created LSCD datasets reflect all of these tasks and thus allow for evaluation of WiC and WSI along with LSCD models \citep[i.a.][]{Schlechtweg2021dwug,Kurtyigit2021discovery,kutuzov2022nordiachange,Zamora2022lscd,Chen2023chiwug}. 

Below, we provide detailed descriptions of each task. Figure~\ref{fig:runningex} sketches the most straightforward and modular approach to LSCD which follows the annotation process of the popular LSCD datasets and explicitly solves each task sequentially: Usages of the target word \textit{backlog} from two time periods are combined into pairs and for each pair a WiC label is predicted. (In this case it is a continuous score, but it could be binary or ordinal.) These are then represented in a graph and clustered. From the clusters, sense probability distributions for the two time periods are derived, which are then compared to measure change, e.g. by the JSD-Shannon Distance \citep{Lin91divergencemeasures} for Graded Change or sense loss and gain for Binary Change. Our benchmark also provides implementations of other approaches to LSCD that either skip some of the tasks or solve them differently from the dataset annotation process. 

\subsection{Word-in-Context}
WiC asks to determine if two words occurring in two text fragments have the same or different meanings. Usually two usages of the same word probably in different grammatical forms are given. 
For example, consider any pair of usages of the word \textit{backlog} in Figure~\ref{fig:runningex}. The WiC task is often framed as a binary classification task. For instance, the WiC~\citep{pilehvar2019wic} and MCL-WiC~\citep{martelli-etal-2021-semeval} datasets contain binary labels and employ accuracy as the main evaluation metric. Alternatively, USim~\citep{Erk13}, SCWS~\citep{huang-etal-2012-improving} and CoSimLex~\citep{Armendariz19} were labeled with non-binary semantic proximity scores and promote a graded formulation of the task. In this formulation, a WiC model shall produce scores that are similar to the human scores, or at least rank the pairs of usages similarly. Spearman's and Pearson's correlation coefficients are employed as evaluation metrics in this case \citep{Spearman1904Association}. More recently, \citet{schlechtweg2025comedi} provided an ordinal formulation of the graded WiC task, evaluating with Krippendorff's $\alpha$ \citep{krippendorff2018content}.

Following the DURel annotation framework for LSCD \citep{schlechtweg2024dureltool}, human annotators solved the graded WiC task, i.e., they annotated the semantic proximity of two usages of the same word on a scale. This provides data for evaluation of WiC models that can serve as a part of LSCD models. In diachronic LSCD datasets, there are pairs of word usages extracted from two documents belonging to distant time periods making usages in these pairs very different orthographically, grammatically, and thematically, even when the target word has the same meaning. This might be challenging for models trained on traditional WiC datasets, which often contain examples from the same time period. Our benchmark helps to analyze how sensitive WiC models are to this shift in time period by comparing their performance on pairs of usages extracted from the old, the new or both corpora.

\subsection{Word Sense Induction}
WSI asks to infer which senses a given target word has based only on its usages in an unlabeled corpus. It is usually framed as a clustering task where a model shall cluster a given set of usages of the same target word probably in different grammatical forms into clusters corresponding to the senses of this word. One way to solve the task is through clustering a weighted graph populated with WiC predictions, cf. the middle part of Figure~\ref{fig:runningex}. Unlike the more popular Word Sense Disambiguation task, in WSI no sense inventory is given to the model and the number of senses of the target word is not known as well. The most widespread formulation of WSI assumes that each usage has one and only one sense, thus, requires hard clustering, i.e. assigning each usage to a single cluster \citep[i.a. SemEval 2010 Task 14, ][]{manandhar-klapaftis-2009-semeval}.

\begin{table*}[t]
\centering
\small
\begin{adjustbox}{width=0.9\textwidth}
\tabcolsep=.09cm
\begin{tabular}{ c | c c c c c c c c c c c c c}            \toprule                           
\textbf{Data set} & \textbf{LGS}  &  $\mathbf{n}$  &  \textbf{N/V/A} &  $\mathbf{|U|}$ &  $\textbf{AN}$ &  \textbf{JUD}  & \textbf{Task} &  $\mathbf{t_1}$ &  $\mathbf{t_2}$ & \textbf{Reference} & \textbf{Version} \\
\midrule  
DWUG & DE & 50 &  32/14/2  & 178 & 8 &  63k  & {\footnotesize WiC, WSI, LSCD (B,G,C)} & {\small 1800--1899} & {\small 1946--1990} & {\small \citet[][]{Schlechtweg2021dwug}} & 3.0.0 \\
{\small DWUG Res.} & DE &15  &10/4/1 & 50&3 &10k & {\small WiC, WSI, LSCD (B,G,C)} & {\small  1800–1899 } & {\small1946–1990 } & {\small \citet[][]{Schlechtweg2024dwugs}}& 1.0.0 \\
DWUG & EN & 46 &  40/6/0  & 191 & 13 &  29k  & {\small WiC, WSI, LSCD (B,G,C)} & {\small 1810--1860} & {\small 1960--2010}& {\small \citet[][]{Schlechtweg2021dwug}} & 3.0.0 \\
{\small DWUG Res.} & EN & 15  & 14/1/0&50 &3 & 7K& {\small WiC, WSI, LSCD (B,G,C)} & {\small1810–1860  } & {\small 1960–2010 } & {\small \citet[][]{Schlechtweg2024dwugs}} & 1.0.0 \\
DWUG & SV & 44 &  32/5/7  & 171 & 13 &  20k & {\small WiC, WSI, LSCD (B,G,C)} & {\small 1790--1830} & {\small 1895--1903}& {\small \citet[][]{Schlechtweg2021dwug}} & 3.0.0 \\
{\small DWUG Res.} & SV & 15 &10/3/2 & 50& 6& 16k& {\small WiC, WSI, LSCD (B,G,C)} & {\small 1790–1830 } & {\small 1895–1903} & {\small \citet[][]{Schlechtweg2024dwugs}} & 1.0.0 \\
DWUG & ES & 100 &  51/24/25 & 40  &  12 & 62k & {\small WiC, WSI, LSCD (B,G,C)} & {\small 1810--1906} & {\small 1994--2020} & {\small \citet[][]{Zamora2022lscd}} & 4.0.2 \\
{\small DiscoWUG} & DE & 75 & 39/16/20 & 49 & 8 & 24k & {\small WiC, WSI, LSCD (B,G,C)}  & {\small 1800--1899} & {\small 1946--1990} & {\small \citet[][]{Kurtyigit2021discovery}} & 2.0.0 \\
 {\small RefWUG } & DE & 22 &  15/1/6  & 19 & 5 &  4k  & {\small WiC, WSI, LSCD (B,G,C)} & {\small 1750--1800} & {\small 1850--1900} & {\small \citet[][]{Schlechtweg2023measurement}} & 1.1.0 \\
{\small NorDiaChange1} & NO & 40 & 40/0/0 & 21 & 3 & 14k & {\small WiC, WSI, LSCD (B,G,C)} & {\small 1929--1965} & {\small 1970--2013} & {\small \citet[][]{kutuzov2022nordiachange}} & 1.0.0 \\
{\small NorDiaChange2} & NO & 40 & 40/0/0 & 21 & 3 & 15k & {\small WiC, WSI, LSCD (B,G,C)} & {\small 1980--1990} & {\small 2012--2019} & {\small \citet[][]{kutuzov2022nordiachange}} & 1.0.0 \\
{\small ChiWUG} & ZH &40  &10/22/8  & 40& 4& 61k & {\small WiC, WSI, LSCD (B,G,C)} & {\small 954-1978} & {\small 1979-2003} & {\small \citet[][]{Chen2023chiwug}} & 1.0.0 \\
{\small DWUG} & IT & 26 & 17/3/6 & 86 & 5 & 5k & {\small WiC, WSI, LSCD (B,G,C)} & {\small 1948-1970 } & {\small 1990-2014} & {\small \citet[][]{cassotti_clicit_2024}} &  3.0.0 \\
\midrule  
DURel  & DE & 22 &  15/1/6 & 104 & 5 &  6k  & {\small WiC, LSCD (C)} & {\small 1750--1800} & {\small 1850--1900} & {\small \citet[][]{Schlechtwegetal18}} & 3.0.0 \\
SURel  & DE & 22 &  19/3/0  & 104 & 4 &  5k  & {\small WiC, LSCD (C)} & {\small general} & {\small domain} & {\small \citet[][]{haettySurel-2019}} & 3.0.0 \\
{\small RuSemShift1} & RU & 71 & 65/6/0 & 119 & 5 & 21k & {\small WiC, LSCD (C)} & {\small 1682--1916} & {\small 1918--1990} & {\small \citet[][]{rodina2020rusemshift}} & 2.0.0 \\
{\small RuSemShift2} & RU & 69 & 57/12/0 & 105 & 5 & 18k & {\small WiC, LSCD (C)} & {\small 1918--1990} & {\small 1991--2016} & {\small \citet[][]{rodina2020rusemshift}} & 2.0.0 \\
{\small RuShiftEval1} & RU & 111 & 111/0/0 & 60 & 3 & 10k & {\small WiC, LSCD (C)} & {\small 1682--1916} & {\small 1918--1990} & {\small \citet[][]{rushifteval2021}} & 2.0.0 \\
{\small RuShiftEval2} & RU & 111 & 111/0/0 & 60 & 3 & 10k & {\small WiC, LSCD (C)} & {\small 1918--1990} & {\small 1991--2016} & {\small \citet[][]{rushifteval2021}} & 2.0.0 \\
{\small RuShiftEval3} & RU & 111 & 111/0/0 & 60 & 3 & 10k & {\small WiC, LSCD (C)} & {\small 1682--1916} & {\small 1991--2016} & {\small \citet[][]{rushifteval2021}} & 2.0.0 \\
\bottomrule                           
\end{tabular}                                        \end{adjustbox}
\caption{Overview datasets. LGS = language, $n$ = no. of target words, N/V/A = no. of nouns/verbs/adjectives, $|U|$ = avg. no. usages per word, AN = no. of annotators, JUD = total no. of judged usage pairs, Task = possible evaluation tasks, $t_1$, $t_2$ = time period 1/2, Reference = data set reference paper, Version = version used for experiments.
}\label{tab:data}
\vspace{-15pt}
\end{table*}

\subsection{Lexical Semantic Change Detection}
LSCD is a general name for several tasks dealing with analysis of different properties of a word related to changes in its meaning over time. 
Usually, in these tasks a list of target words are given and two time periods are specified, an old and a new one. 
Each time period is represented by an unlabeled corpus or a pre-selected set of usages.

The Binary Change task~\citep{schlechtweg-etal-2020-semeval, Zamora2022lscd} asks if the set of senses of a given word is the same for two time periods, i.e., whether any senses were lost or gained, or not. It assumes that word meaning in a particular time period can be described as a set of discrete and mutually exclusive senses observed in the corresponding corpus. This task can be viewed as a task of binary classification of words. 
The Graded Change (or JSD) task~\citep{schlechtweg-etal-2020-semeval, Zamora2022lscd} has the same assumptions about word meaning, but instead of binary classification it requires ranking a given list of words according to changes in their sense frequency distributions. The rank of a word is determined by the Jensen–Shannon Distance between two probability distributions over word senses $P(sense|w, t_{old})$ and  $P(sense|w, t_{new})$, one for the older time period and another for the newer one~\citep{schlechtweg-etal-2020-semeval, Zamora2022lscd}.
The Binary and Graded Change tasks are illustrated in the right part of Figure~\ref{fig:runningex}. A common way to derive sense probability distributions is through WiC-based WSI, middle part of Figure~\ref{fig:runningex}.

Finally, there is also the COMPARE task~\citep{Schlechtwegetal18,rushifteval2021,Zamora2022lscd}, which requires ranking a given set of words according to the average proximity (WiC score) between old and new usages of each word, i.e., the average cross-period proximity. This task avoids clustering and can be seen as a simple approximation of the Graded Change task.

\section{Datasets}
\label{sec:datasets}

Table \ref{tab:data} shows all datasets currently integrated into the benchmark. All datasets have in common that they are based on human WiC judgments of word usage pairs on the ordinal DURel scale from 1 to 4, where 1 means semantically unrelated and 4 means identical \citep[][]{Schlechtwegetal18}. They also share the use of diachronic data.\footnote{With the exception of SURel comparing usages from different domains rather than time periods \citep[][]{haettySurel-2019}.} 

The datasets then fall into two main categories: (i) Datasets representing annotated judgments in a sparsely connected graph (Word Usage Graph, find an example in Figure \ref{fig:plane}, in Appendix \ref{sec:annotated}), clustering these with a variation of correlation clustering \citep{Bansal04,Schlechtweg2021dwug}, and deriving LSC labels by comparing the two time-specific sense frequency distributions \citep{schlechtweg-etal-2020-semeval}, as illustrated in Figure~\ref{fig:runningex}. These datasets, displayed in the upper part of Table \ref{tab:data}, apply the full lexicographic process and thus allow for full evaluation on all tasks mentioned in Section \ref{sec:tasks}. (ii) Datasets skipping the clustering step and hence only allowing for evaluation on the WiC and COMPARE tasks. These are shown in the lower part of Table \ref{tab:data}.
Apart from the difference in tasks they support, the datasets have strongly varying properties in terms of language, number of target words, PoS distribution, number of usages per target word and number of human judgments. We integrate provided preprocessed usages (tokenization, lemmatization, normalization, PoS) as preprocessing options together with general strategies such as target word substitution by the lemma. There is a range of further data which could be integrated into the benchmark with minor adjustments such as WiC or Raw-C \citep{pilehvar2019wic,trott-bergen-2021-raw}.

\begin{figure*}[t]
   \centering
       \includegraphics[trim=0cm 14cm 0cm 6cm, clip, width=.95\textwidth]{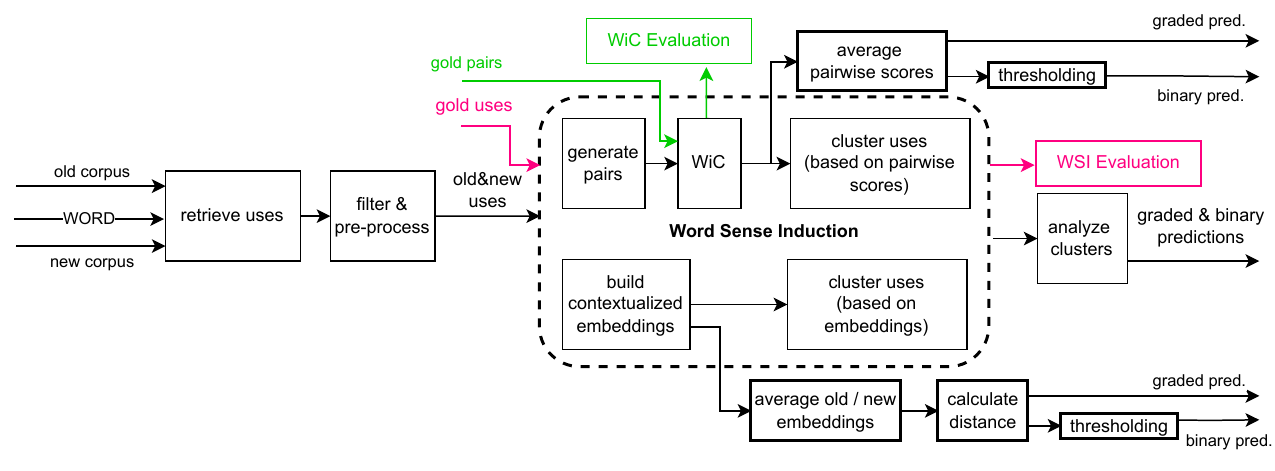} 
\vspace{-10pt}
 \caption{Token-based LSCD pipelines and their evaluation.}
 \label{fig:models}
\vspace{-15pt}
\end{figure*}

\section{Evaluation Procedures}
\label{sec:eval}
Figure~\ref{fig:models} shows the structure of our benchmark. It summarizes token-based approaches to LSCD and shows how their components and whole pipelines can be evaluated using our benchmark. The benchmark reflects the main LSCD approach illustrated in Figure~\ref{fig:runningex}, but also allows for simpler modeling, e.g. by skipping the clustering step.

The central part shows the WSI-based approach to LSCD. It relies on WSI methods which cluster word usages based either on their contextualized embeddings or pairwise similarities between them calculated by a WiC model. If a WiC model is involved, we can evaluate it separately on all datasets containing human-labeled pairs of word usages by feeding these pairs and comparing the model predictions with the human labels. Spearman's and Pearson's correlation coefficients that compare rankings or scores predicted by humans and the WiC model are employed as metrics for the WiC task. 
A WSI method can also be evaluated as a whole by running it on a set of gold usages of each word, i.e. usages that have sense labels obtained either directly from human annotators, or by clustering Word Usage Graphs. Clusterings obtained with the WSI method are compared against sense labels using the Adjusted Rand Index~\citep{Hubert1985} as a the main metric.
Finally, we can evaluate the whole LSCD pipeline using the standard LSCD metrics, i.e. F1-score for the binary classification tasks or Spearman's correlation with the gold word ranking for the JSD and COMPARE tasks. 

We introduce standard splits for each dataset on the lemma level, i.e., certain target words are assigned to train/dev/test. We further provide possibility to evaluate on the previously introduced standard split from the CoMeDi shared task \citep{schlechtweg2025comedi}. Find an overview of the number of target words in this split in Appendix \ref{sec:comedisplit}.

\section{Models}
\label{sec:models}

The most straightforward modeling approach for LSCD models is to follow the 3-level annotation approach of WUGs described in Section \ref{sec:tasks} and displayed in Figure~\ref{fig:runningex}. Given a target word, a basic model retrieves uses of this word from an \textit{old} and a \textit{new} corpus, then clusters them in order to infer word senses, and finally analyzes the obtained clusters to make conclusions about changes in word senses between two time periods. This approach is appealing because if word senses are inferred correctly, then an exact description of how word senses changed, as well as the exact predictions for all LSCD tasks are easy to obtain. Also the procedure basically automates the annotation procedure of various LSCD datasets, which is a reasonable way of getting predictions that correspond well to the ground truth. The benchmark implementation as depicted in Figure \ref{fig:models} thus follows this basic structure. The inputs and the outputs of each component are specified in round brackets, while the hyperparameters are underlined in the corresponding descriptions. The implemented components on each of these levels will be described below. It is important to note that current SOTA models for Graded Change \citep{giulianelli-etal-2020-analysing,Arefyev202116,Cassotti2023Xllexeme} skip the clustering step and model Graded Change directly from WiC predictions of proximity between word usages or their underlying vectors by aggregating these time-wise (see below).

\begin{figure*}[t]
    \begin{subfigure}{0.5\textwidth}
\includegraphics[width=\linewidth,trim={0 0 0 0},clip]{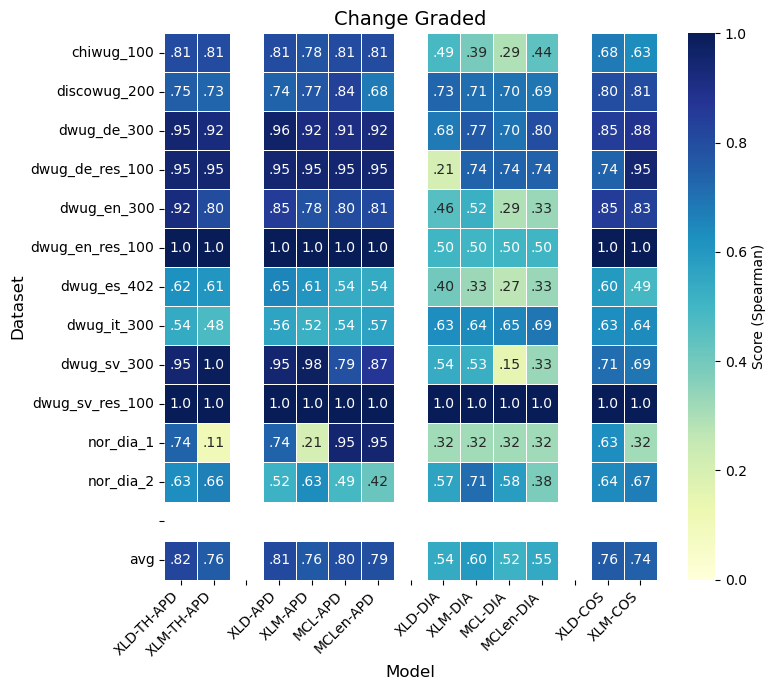}
    \end{subfigure}
    \begin{subfigure}{0.5\textwidth}
\includegraphics[width=\linewidth,trim={0 0 0 0},clip]{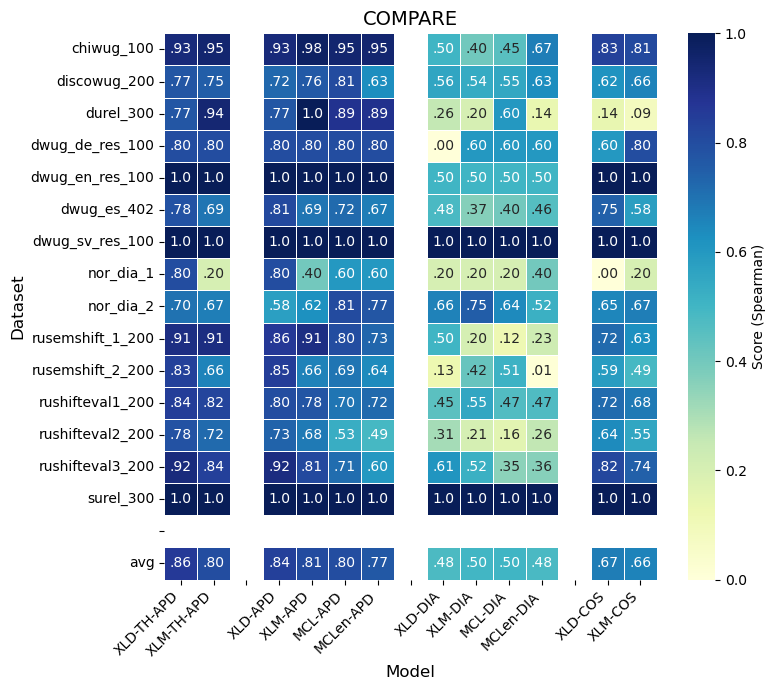}
    \end{subfigure}
    \caption{Result overview on Graded Change (left) and COMPARE score (right). XLD = XL-DURel, XLM = XL-LEXEME, MCL/MCLen =  DeepMistake checkpoints, TH = thresholded, APD = Average Pairwise Distance, DIA = DiaSense, COS = Cosine distance between average embeddings.}\label{fig:results-lscd}
\vspace{-15pt}
\end{figure*}
\paragraph{Retrieve uses} (word, corpus → uses). Given a corpus and a target word, this component retrieves all uses of the target word in all of its grammatical forms from the given corpus. Optionally, \underline{N} uses may be sampled if there are more than that in order to reduce the following computations.

For intrinsic evaluation, instead of retrieving uses, golden uses may be taken, i.e., the uses which were shown to the annotators during dataset construction. This eliminates possible disagreements due to some rare senses of the target word sampled during the dataset annotation procedure but not sampled by the model during evaluation, or vice versa.

\paragraph{Build contextualized embeddings} (uses → embeddings). We employ BERT or XLM-R masked language \textit{model} to obtain the contextualized representations of the given target word in the given text fragment \citep{devlin-etal-2019-bert,conneau2019unsupervised}.\footnote{In principle, the benchmark supports all model checkpoints from huggingface.} 
The simplest and most popular option is taking the outputs of the last Transformer layer (before the MLM head) on the positions of the target word and average those outputs (mean pooling). We can apply other \textit{subword poolings}, max pooling or first pooling (take the outputs from the first subword). A more general implementation combines the outputs from several \textit{layers}. Again, the simplest option is just averaging them.\footnote{Check \citet{vulic-etal-2020-probing} for a detailed study on averaging layers for lexical tasks.} However, the experiments with BERT without fine-tuning presented in~\citet{devlin-etal-2019-bert} for the NER task suggest that it may be better to concatenate the outputs of the last four layers instead of averaging them. Thus, the \textit{layer aggregation function} is a hyperparameter and we select its value among averaging and concatenation. Additionally, the benchmark supports XL-LEXEME, a WiC-fine-tuned version of XLM-R \citep{Cassotti2023Xllexeme} and the similar XL-DURel which is instead fine-tuned for ordinal WiC \citep{Yadav2025xldurel}.

\paragraph{Generate pairs} (uses → pairs of uses). Pairs of uses of the specified \textit{type} are generated. For the COMPARE type, each pair contains a use from the old corpus and a use from the new corpus. This type is ideal for the COMPARE task in which one needs to estimate the average proximity between old and new usages. If type is ALL, then all possible pairs are generated. This is useful to provide more information for the following clustering step.

\paragraph{WiC} (pairs of uses → pairwise scores). We consider two different approaches to the WiC task. The first approach builds the contextualized embeddings for all uses with the aforementioned component, it inherits all the corresponding \textit{hyperparameters}. Then, it employs one of the \textit{distance functions} to calculate distances between the contextualized embeddings of two uses in each pair. The euclidean, manhattan and cosine distances are currently supported.

The second approach employs a binary classifier implemented as a neural network that jointly processes a pair of word usages and returns the probability that the meaning is the same. We treat this probability as the proximity between word occurrences. We experimented with the DeepMistake system\footnote{DeepMistake is a family of WiC models. It includes models fine-tuned on the data from the LSCD shared tasks in Russian or Spanish, and more general WiC models trained on general-purpose multilingual WiC datasets which we employ in our study. Specifically, we selected the MCL and MCLen models trained on the whole or the English part of the MCL-WiC dataset respectively. We further compare them to the language-specific models in Appendix~\ref{sec:deepmistake}.} which had achieved the 2nd best result in the RuShiftEval and LSCDiscovery shared tasks~\citep{Arefyev202116,deepmistake-lcsdiscovery}, and also with the method from \citep{Yadav2025xldurel,schlechtweg2025comedi} which has shown good performance on the recent ordinal WiC shared task data. 

We further provide the possibility of discretizing graded WiC predictions at specified thresholds. By default, the thresholds from \citet{Yadav2025xldurel} are used.

\paragraph{Clustering} (pairwise scores → clusters or contextualized embeddings → clusters). The goal of this step is to discover all senses of the target word occurring in the old, or the new corpus, or both of them, i.e., solve the WSI task for all uses retrieved from both corpora. Clusters are inferred simultaneously on all usages from old and new corpus. The clustering algorithms that can accept a matrix of similarities or distances between objects instead of object vectors can be applied to both types of inputs to the clustering component (pairwise scores, embeddings). However, some clustering algorithms require raw object vectors and can be applied to the contextualized embeddings, but not to the pairwise scores. For instance, K-Means calculates cluster centroids and needs object vectors to do that. The benchmark currently supports correlation clustering \citep{Bansal04} operating on pairwise scores. We chose this algorithm as most gold annotations were clustered with this approach \citep[e.g.][]{schlechtweg-etal-2020-semeval}.

\paragraph{Cluster measures} (clusters → LSCD predictions). To solve the Binary Change task, after WSI we search for those clusters that contain only new or only old examples. Those clusters are viewed as novel or lost senses correspondingly, and the binary labels are predicted accordingly. Alternatively, we search for the clusters with at least \textit{M} new examples and at most \textit{K} old examples or vice versa in order to align with the annotation procedure of some LSCD datasets. 
To solve the JSD task, for old and new uses separately we estimate the probability distribution over senses of the target word as the proportions of its uses present in each cluster. Then the JSD between the two probability distributions is calculated. 

\begin{figure}
    \centering    
    \includegraphics[width=0.7\linewidth]{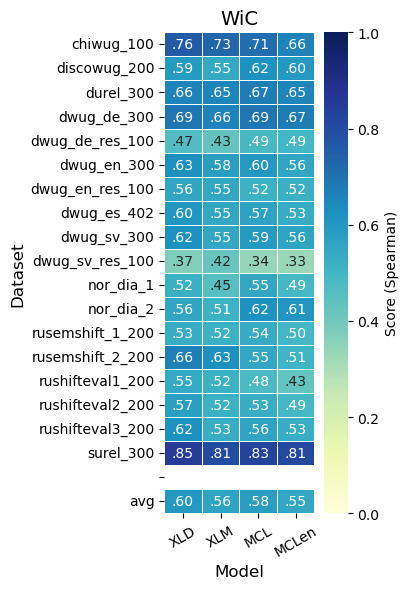}
    \caption{Result overview on WiC.}
    \label{fig:results-wic}
\vspace{-20pt} 
\end{figure}

\paragraph{Aggregate measures} (pairwise scores → LSCD predictions or contextualized embeddings → LSCD predictions). These skip the clustering step by aggregating WiC predictions directly. The most successful of these is \textit{Average Pairwise Distance (APD)}, which simply averages the distances or similarities between word usages from different time periods (COMPARE pairs) returned by some WiC model \citep{kutuzov-giulianelli-2020-uiouva}. This follows the calculation of the gold COMPARE scores, which is the average of human pairwise annotations.
APD is sensitive to the polysemy or variation of the target word, which can lead to wrong predictions. For this, measures normalizing APD by a polysemy term were proposed, such as \textit{DiaSense} \citep{beck-2020-diasense}. \textit{COS}, in contrast, averages the contextualized embeddings of all old and all new uses separately producing two aggregated embeddings of the target word for each of of the two time periods. Then, it calculates the distance between those two embeddings. It was previously introduced under the names of PRT~\citep{kutuzov-giulianelli-2020-uiouva} and COS~\citep{Laicher2021explaining}, cf. also \citet{martinc-etal-2020-leveraging}.
This distance is used as the prediction for both graded tasks (Graded Change and COMPARE).

\section{Experiments}
\label{sec:experiments}

We now use the benchmark to perform a number of experiments. We focus on Graded Change and COMPARE detection as these are the most widely approached LSCD tasks \citep{rushifteval2021,Periti2024systematic}. Note that not all datasets provide both evaluation scores (see Table \ref{tab:data}). All experiments are performed on the CoMeDi test split described in Section \ref{tab:data}. We cannot test all models and configurations, thus we focus on SOTA models using aggregate change measures without clustering and design experiments to answer open research questions.

\paragraph{Which WiC model and which aggregate measure gives SOTA performance?} According to recent studies \citep{Zamora2022lscd,Periti2024systematic,zamorareina2025lscd}, DeepMistake and XL-LEXEME combined with APD compete for the SOTA on Graded Change and COMPARE detection. Previous studies have not directly compared these models though, or only on very limited data \citep{zamorareina2025lscd}. Hence, we perform a direct comparison guaranteeing a fair evaluation setup. We also include the recently published XL-DURel \citep{Yadav2025xldurel}, which has been shown to improve upon both models on ordinal WiC. In addition to the canonical datasets, we include a number of recently published datasets which have never been used for model evaluation: DWUG DE/EN/SV V3.0.0, DWUG DE/EN/SV resampled and DWUG IT. This is the most thorough model comparison done so far in the field of LSCD.

\begin{figure}
    \centering    
    \includegraphics[width=0.8\linewidth]{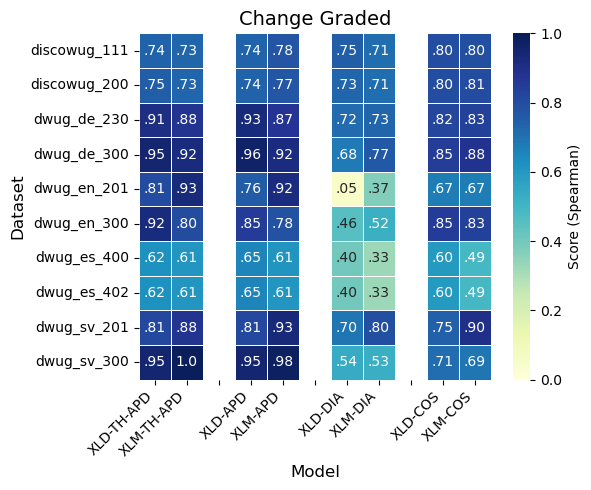}
    \caption{Result overview on dataset versions for bi-encoder models.}
    \label{fig:results-version}
\vspace{-20pt}
\end{figure}

Figure \ref{fig:results-lscd} shows the results across the three SOTA WiC models, each combined with the common aggregate change measures APD, COS and DiaSense (see Section \ref{sec:models}). Both, for Graded Change and COMPARE, clearly APD dominates which confirms previous results. However, there are slight advantages for the recently published XL-DURel model suggesting a new SOTA.

\paragraph{Does WiC prediction discretization improve results?} All above-reported measures are based on aggregation of graded WiC predictions (either cosine similarity or same-sense probability). However, all WUG datasets are annotated on an ordinal (discrete) scale (see Section \ref{sec:datasets}). Exactly predicting these ordinal values can be done by thresholding the graded predictions \citep{Choppa2025comedi}. We hypothesize that discretizing WiC predictions to resemble human annotations helps for LSCD as ordinal judgments were used for ground-truth construction. Hence, in Figure \ref{fig:results-lscd} we report two models with APD and thresholding (XLD/XLM-TH-APD). Threshold parameters were taken from \citet{Yadav2025xldurel}. As we see, for XL-DURel, thresholding slightly helps, further pushing the SOTA, while for XL-LEXEME it has no effect or slightly hurts.

\paragraph{Which model gives SOTA on diachronic WiC? Does WiC determine LSCD?} A recent shared task has compared DeepMistake and XL-LEXEME for ordinal WiC showing a slight advantage for DeepMistake \citep{schlechtweg2025comedi}. 
XL-DURel has further improved upon both models. Performance on the WiC task is a strongly influential factor for LSCD \citep{Arefyev2021Deep}. We compare model performance on these two levels to gain an understanding how strongly WiC determines LSCD performance. Figure \ref{fig:results-wic} compares WiC models on the ordinal WiC task. The MCL checkpoint of DeepMistake does dominate XL-LEXEME confirming previous results, but both are outperformed by XL-DURel. Now compare this to the APD columns in Figure \ref{fig:results-lscd}. Overall, WiC performance determines LSCD performance. Consider e.g. the dominance of XL-DURel on DWUG EN with Graded Change. However, there are notable exceptions such as DWUG SV where XL-DURel dominates on WiC but not LSCD. This shows that purely improving WiC ranking does not guarantee better LSCD performance. Possibly, the score distribution plays an important role when averaging values for aggregate measures.

\paragraph{Are model performances reproducible with more reliable data? What is the performance development on incrementally annotated datasets?} Many datasets have been annotated in incremental rounds of annotation \citep{Schlechtweg2021dwug,Schlechtweg2024dwugs}. Later rounds are supposed to yield higher data quality as graphs are more richly annotated. \citet{Schlechtweg2024dwugs} suggest that previous model comparisons done on older dataset versions/less rounds should be repeated with the more reliable data (last round). We are the first to investigate model performance with the latest dataset versions. This will also allow us to investigate the impact of annotation rounds on performance. Figure \ref{fig:results-version} shows the performance of a selection of models on consecutive versions. For the top models, we can see that performance tends to increase with later versions, which is expected as quality should increase. However, there are cases where model performance strongly drops for newer versions, such as XLM-APD on DWUG EN. Further, the relative performance of models can strongly change depending on version, e.g. XLM-APD outperforms XLD-APD on DWUG EN for an older version while it is the other way around for the newer, more reliable version. This shows the risks of using unreliable ground-truth data and illustrates the need for the creation of benchmarks such as ours and continuous model reevaluation on additional and improved datasets.

\begin{figure}[t]
\centering
    \begin{subfigure}{0.2\textwidth}
\includegraphics[width=\linewidth,trim={0 0 2.45cm 0},clip]{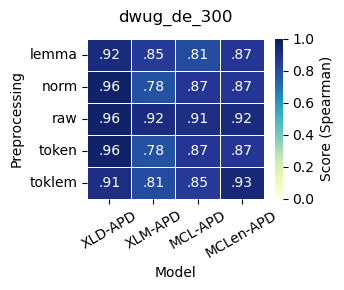}
    \end{subfigure}
    \begin{subfigure}{0.19\textwidth}
\includegraphics[width=\linewidth,trim={2.8cm 0 0 0},clip]{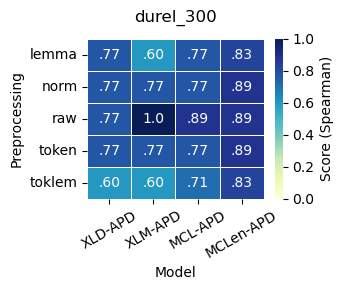}
    \end{subfigure}
    \caption{Result selection on preprocessing. lemma = lemmatization, norm = normalization, raw = no preprocessing, token = tokenization, toklem = tokenization with target word substituted by lemma.}\label{fig:results-preprocessing}
\vspace{-15pt}
\end{figure}

\paragraph{What is the impact of spelling variation on performance?} \citet{Laicher2020} show that historical spelling variations can have a strong influence on BERT-based model performance. However, it is not clear how much this influences SOTA models. Hence, we test the influence of spelling normalization and lemmatization on two German datasets containing historical spelling variants in Figure \ref{fig:results-preprocessing}. We can see that applying no preprocessing (raw) gives top performance for all models, with one exception (MCLen with toklem). This suggests that current base embedders are quite robust and do not need additional preprocessing. 

\paragraph{Relation to previous work.} We did not evaluate models on the full set of target words from each dataset, but only on the CoMeDi test split, as described in Section \ref{sec:eval}. This enabled us to test models such as XL-DURel, which was fine-tuned on the use-pair-level data of many LSCD datasets making it impossible to be tested on the full data. This also means that most previous results are not directly comparable to ours as they were obtained on the full (or almost full) datasets. On the WiC-level, the results from \citet{Yadav2025xldurel} are weakly comparable as they were obtained on the same split, but with a stronger cleaning strategy and data concatenated per language. This leads to lower top results in our evaluation on German data (except for SURel), English and Spanish, higher results for Chinese and Norwegian, and mixed results for Swedish and Russian. Similarly, on the LSCD-level, the results from \citet{Periti2024systematic} are weakly comparable as they were obtained on the full datasets including the split we evaluated on. Our top results are higher for Chinese, German, English, Swedish, and similar for Russian, Spanish and Swedish. In the future, it will be essential (and also very simple) to compare models on the full data to make the results more comparable to previous work.

\paragraph{Results on concatenated data.} We experimented with evaluating models on different concatenations of datasets and noticed that some model performance differences vanish in this setup. In a further analysis of model predictions, we realized that at least part of the reason for this effect are prediction outliers in individual datasets which deteriorate model performance much more in larger concatenations of data. While concatenating datasets is a good way to overcome data sparsity, we are unsure about the validity of this approach as it assumes that datasets were annotated consistently and that dataset-specific properties such as the size of the annotated graphs do not introduce biases that would make change scores incomparable. We leave more research in this direction to future work.

\section{Conclusion}
In this work, we have presented a new benchmark for evaluation of token-based LSCD models. The procedures are implemented that can evaluate both whole LSCD solutions and their separate components that solve the WiC and WSI subtasks. A variety of LSCD datasets are integrated in the benchmark allowing thorough evaluation on various languages and diverse historical epochs. We used the benchmark to perform a number of experiments with recent models setting a new SOTA in LSCD and providing a better general understanding on LSCD model evaluation and improvement.  

Our main findings can be summarized as follows: (i) Recent WiC models optimized for ranking use pairs on an ordinal scale perform competitively or better than WiC models optimized on binary WiC data. (ii) Discretizing graded WiC predictions to ordinal, human-like values helps for ordinal WiC models. (iii) Overall, WiC performance determines LSCD performance. (iv) Using less reliable, older versions of datasets can lead to false conclusions on optimal model choice. (v) Current base embedder models are quite robust against spelling variation. Overall, our results suggest a kind of trivial, but rarely acknowledged trend in LSCD: More closely modelling the human annotation process leads to better performance.  

\section*{Limitations}
In our evaluation, we did not evaluate any cluster-based models although these have a large potential for high performance \cite{Zamora2022lscd,Schlechtweg2024sense}. However, the current SOTA does not build on clustering. Hence, we focused our evaluation to give a more concise overview and leave the comparison to cluster-based models to future work. We have also used seemingly non-optimal DeepMistake model training checkpoints, likely underestimating its performance. Another problem is the small size of the CoMeDi test split, potentially leading to strong performance variations through sampling error.

In the future, we want to use the benchmark to compare clustering models including a comparison on full datasets. We also want to add more datasets such as the Slovenian or the Japanese one \citep{pranjic2025trackingsemanticchangeslovene,ling-etal-2023-construction,Baldissin2022diawug}, provided they are publicly available in the correct format. This will have the advantage that we can test all models on the full datasets as they have not been part of the training data. Further, we want to add more recently developed semantic change metrics \citep{pranjic2025trackingsemanticchangeslovene,goworek-dubossarsky-2026-rethinking} and base embedders optimized for ordinal WiC prediction \citep{Mujko2026thesis}. On the long term, we also see the possibility of integrating the benchmark with more fine-grained LSCD datasets providing semantic change types \citep{cassotti-etal-2024-using,Whaley2025thesis} or a definition generation component \citep{fedorova2024axolotl}.

While our aim is to standardize LSCD model evaluation, we do not aim to delegitimize other evaluation approaches or variations of the one used here. In our benchmark, we try to capture as much heterogeneity as we can within the current main LSCD paradigm, but approaches focusing on types of change \citep{cassotti-etal-2024-using,Whaley2025thesis} or detecting changes against a dictionary \citep{fedorova2024axolotl,Blessing2026apodictus} are equally valid and should be pursued further, in our opinion.

\section*{Acknowledgments}
Dominik Schlechtweg and Sachin Yadav have been funded by the research program `Change is Key!' supported by Riksbankens Jubileumsfond (under reference number M21-0021). Nikolay Arefyev has received funding from the European Union’s Horizon Europe research and innovation program under Grant agreement No 101070350 (HPLT).  Thanks to Andres Cabero, Kuan-Yu Lin and Arshan Seyed Dalili for contributing code to the repository. Thanks to Shafqat Mumtaz Virk for contributing to an earlier version of this paper.

\bibliography{additional-references, bibliography-self, Bibliography-general, bibliography-supervision-self}

\clearpage
\appendix

\section{CoMeDi Data Split}
\label{sec:comedisplit}

\begin{figure}[t]
\centering
    \includegraphics[width=0.5\textwidth]{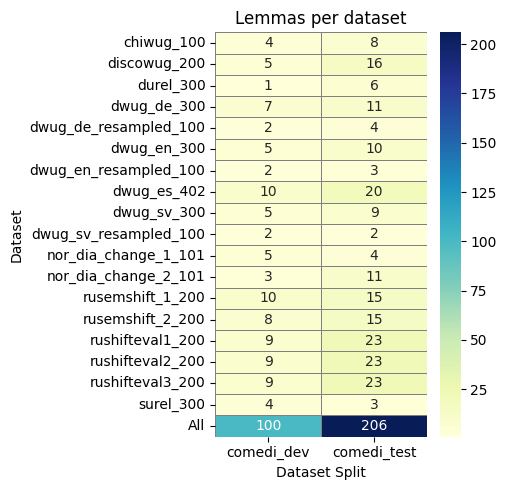}
    \caption{Number of target words in the CoMeDi split per dataset.}
    \label{fig:comedisplit}
\vspace{-20pt}
\end{figure}

Find an overview of the number of target words in the dev and test portions of the CoMeDi data split in Figure \ref{fig:comedisplit}.

\section{Specialized DeepMistake Models}
\label{sec:deepmistake}

For our main experiments we have selected the DeepMistake models that were trained on the general-purpose multilingual WiC dataset MCL-WiC~\citep{martelli-etal-2021-semeval}. However, the best models from~\citep{Arefyev202116} and \citep{deepmistake-lcsdiscovery} were further fine-tuned on the training subsets of the corresponding LSCD shared tasks. 

Table~\ref{tab:dm-cmp-wic} compares the MCL model and the same model further fine-tuned on the RuSemShift dataset. We observe the largest improvements on the Russian datasets, and also surprisingly on the Spanish dataset in exchange to a big loss of performance on DURel. On other datasets the difference is 0.02 or smaller.

Table~\ref{tab:dm-cmp-lscd} further studies how the difference in these two WiC models affect their performance in LSCD. In these experiment we used the APD approach to LSCD, and averaged the WiC predictions for the COMPARE pairs left from the previous experiment, i.e. we used only the golden COMPARE pairs due to the limited computational budget. Similarly to the WiC task, we observe a big boost in LSCD performance on the Russian and Spanish datasets, but now the margins are larger. There are small or no changes on most other datasets. The exception this time is the resampled Swedish dataset where the ordering was mirrored, but this is likely due to chance because there are only 2 words in the CoMeDi test part of this dataset.

To summarize, the results in this section once again confirm that fine-tuning on the target LSCD datasets can hugely boost the performance of WiC models for the LSCD task. Though the XL-DURel which was trained on all LSCD datasets generally outperforms other models across a wide range of LSCD datasets and languages, when focusing on a particular language training a specialized DeepMistake models for this language may give competitive results.

\pgfplotstableread[col sep=comma]{
dataset,MCL,WICRSS,improvement

chiwug\_100,0.71,0.69,-0.02
discowug\_200,0.62,0.61,-0.01
durel\_300,0.67,0.61,-0.06
dwug\_de\_300,0.69,0.69,0.00
dwug\_de\_resampled\_100,0.49,0.50,0.01
dwug\_en\_300,0.60,0.59,-0.02
dwug\_en\_resampled\_100,0.52,0.49,-0.02
dwug\_es\_402,0.57,0.61,0.04
dwug\_sv\_300,0.59,0.59,0.00
dwug\_sv\_resampled\_100,0.34,0.36,0.02
nor\_dia\_change\_1\_101,0.55,0.56,0.02
nor\_dia\_change\_2\_101,0.62,0.63,0.01
rushifteval1\_200,0.48,0.54,0.06
rushifteval2\_200,0.53,0.57,0.04
rushifteval3\_200,0.56,0.60,0.04
surel\_300,0.83,0.84,0.01                     
}\datatable

\begin{table}[t]
\scriptsize 
\centering
\begingroup
\catcode`\_=12 
\pgfplotstabletypeset[
  columns={dataset,MCL,WICRSS,improvement},
  every head row/.style={before row=\toprule, after row=\midrule},
  every last row/.style={after row=\bottomrule},
  columns/dataset/.style={
    column name=\textbf{Dataset},
    string type,
    column type={>{\ttfamily\detokenize{}}c}
  },
  columns/MCL/.style={column name=\textbf{MCL}, fixed, precision=2,zerofill},
  columns/WICRSS/.style={column name=\textbf{MCL$\rightarrow$RSS}, fixed, precision=2,zerofill},
  columns/improvement/.style={
    column name=$\Delta$,
    fixed, precision=2,
    postproc cell content/.code={
      \pgfmathsetmacro{\cellvalue}{##1}
      \ifdim\cellvalue pt<0pt
        \pgfmathsetmacro{\intensity}{-\cellvalue*500}
        \edef\temp{\noexpand\cellcolor{red!\intensity}\pgfmathprintnumber[fixed,precision=2,zerofill]{##1}}
      \else\ifdim\cellvalue pt>0pt
        \pgfmathsetmacro{\intensity}{\cellvalue*500}
        \edef\temp{\noexpand\cellcolor{green!\intensity}\pgfmathprintnumber[fixed,precision=2,zerofill]{##1}}
      \else
        \edef\temp{\noexpand\cellcolor{white}\pgfmathprintnumber[fixed,precision=2,zerofill]{##1}}
      \fi\fi
      \pgfkeyslet{/pgfplots/table/@cell content}\temp
    }
  }
]\datatable
\endgroup
\caption{Comparison of the general and specialized DeepMistake models on the WiC task. The results on RuSemShift are skipped because it was used as training and development data for the specialized model. Spearman's correlation with the human annotations when sorting pairs of usages is reported.}
\label{tab:dm-cmp-wic}
\end{table}

\pgfplotstableread[col sep=comma]{
dataset,MCL,WICRSS,improvement
chiwug\_100,0.95,0.95,0.00
discowug\_200,0.81,0.73,-0.08
durel\_300,0.94,0.94,0.00
dwug\_de\_300,0.95,0.95,-0.01
dwug\_de\_resampled\_100,0.80,0.80,0.00
dwug\_en\_300,0.95,0.90,-0.05
dwug\_en\_resampled\_100,1.00,1.00,0.00
dwug\_es\_402,0.74,0.88,0.15
dwug\_sv\_300,0.90,0.88,-0.02
dwug\_sv\_resampled\_100,-1.00,1.00,2.00
nor\_dia\_change\_1\_101,0.60,0.60,0.00
nor\_dia\_change\_2\_101,0.81,0.78,-0.03
rushifteval1\_200,0.66,0.86,0.20
rushifteval2\_200,0.53,0.74,0.21
rushifteval3\_200,0.70,0.82,0.12      
surel\_300,1.00,1.00,0.00                     

}\datatable

\begin{table}[t]
\scriptsize 
\centering
\begingroup
\catcode`\_=12 
\pgfplotstabletypeset[
  columns={dataset,MCL,WICRSS,improvement},
  every head row/.style={before row=\toprule, after row=\midrule},
  every last row/.style={after row=\bottomrule},
  columns/dataset/.style={
    column name=\textbf{Dataset},
    string type,
    column type={>{\ttfamily\detokenize{}}c}
  },
  columns/MCL/.style={column name=\textbf{MCL}, fixed, precision=2,zerofill},
  columns/WICRSS/.style={column name=\textbf{MCL$\rightarrow$RSS}, fixed, precision=2,zerofill},
  columns/improvement/.style={
    column name=$\Delta$,
    fixed, precision=2,
    postproc cell content/.code={
      \pgfmathsetmacro{\cellvalue}{##1}
      \ifdim\cellvalue pt<0pt
        \pgfmathsetmacro{\intensity}{-\cellvalue*200}
        \edef\temp{\noexpand\cellcolor{red!\intensity}\pgfmathprintnumber[fixed,precision=2,zerofill]{##1}}
      \else\ifdim\cellvalue pt>0pt
        \pgfmathsetmacro{\intensity}{min(100,\cellvalue*200)}
        \edef\temp{\noexpand\cellcolor{green!\intensity}\pgfmathprintnumber[fixed,precision=2,zerofill]{##1}}
      \else
        \edef\temp{\noexpand\cellcolor{white}\pgfmathprintnumber[fixed,precision=2,zerofill]{##1}}
      \fi\fi
      \pgfkeyslet{/pgfplots/table/@cell content}\temp
    }
  }
]\datatable
\endgroup
\caption{Comparison of the general and specialized DeepMistake models on the COMPARE task. Spearman's correlation with the ground truth ranking of words is reported.}
\label{tab:dm-cmp-lscd}
\vspace{-20pt}
\end{table}

\section{Annotated Data Example}
\label{sec:annotated}

\begin{figure*}[t]
    \begin{subfigure}{0.33\textwidth}
\frame {        \includegraphics[width=\linewidth]{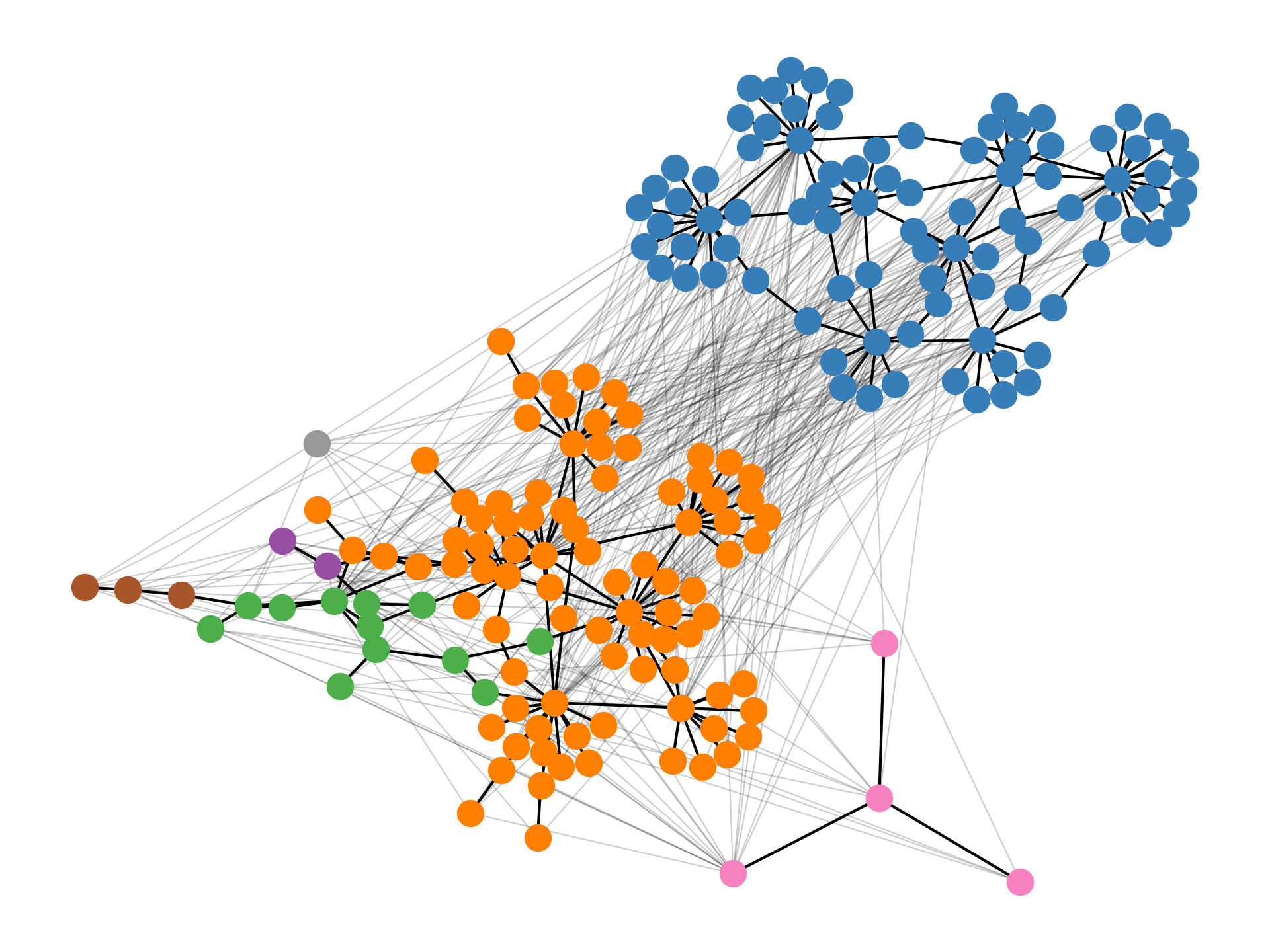}}
%        \caption*{$G$}
    \end{subfigure}
    \begin{subfigure}{0.33\textwidth}
\frame{        \includegraphics[width=\linewidth]{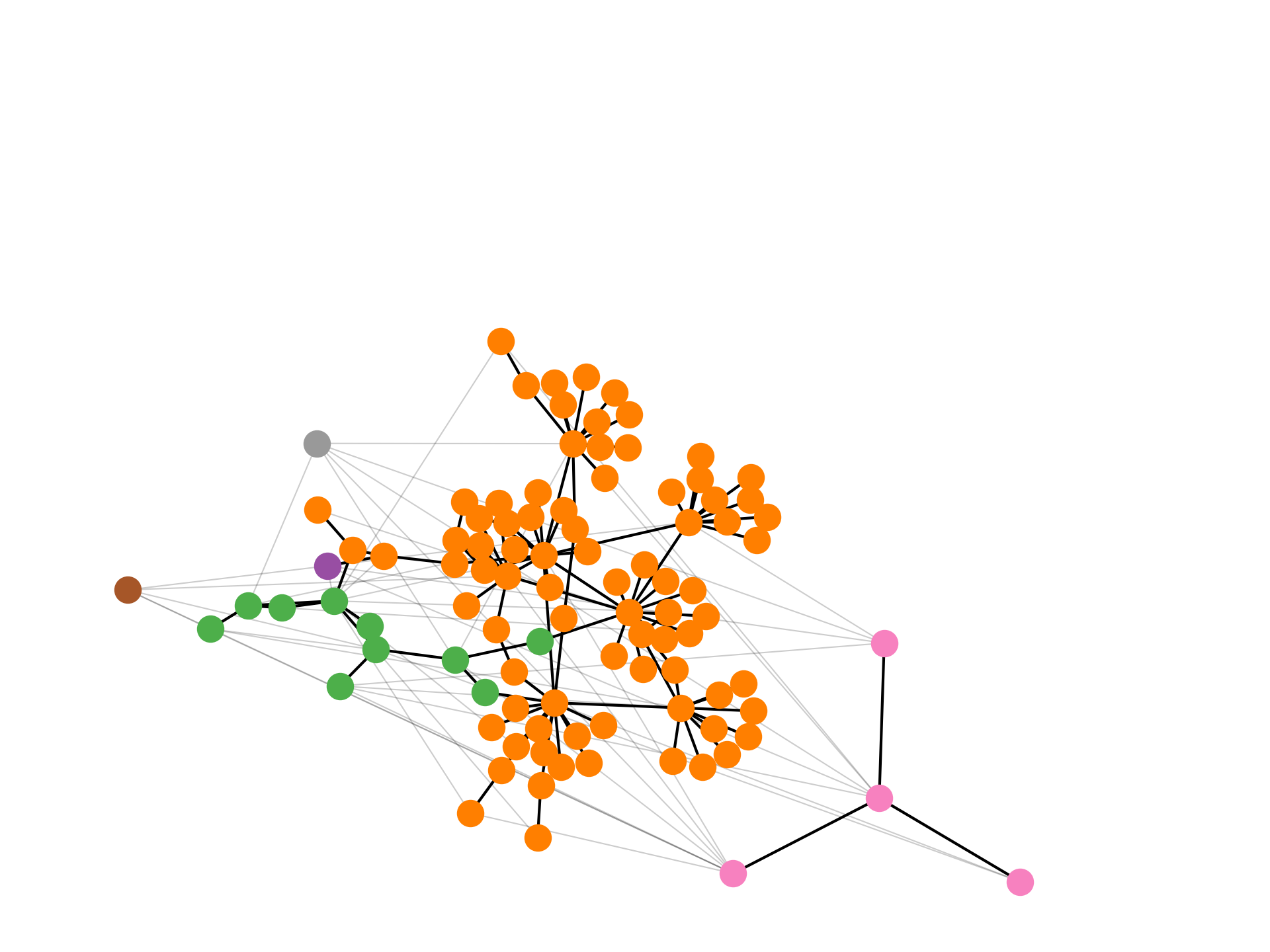}}
%        \caption*{$G_1$}
    \end{subfigure}
    \begin{subfigure}{0.33\textwidth}
\frame{        \includegraphics[width=\linewidth]{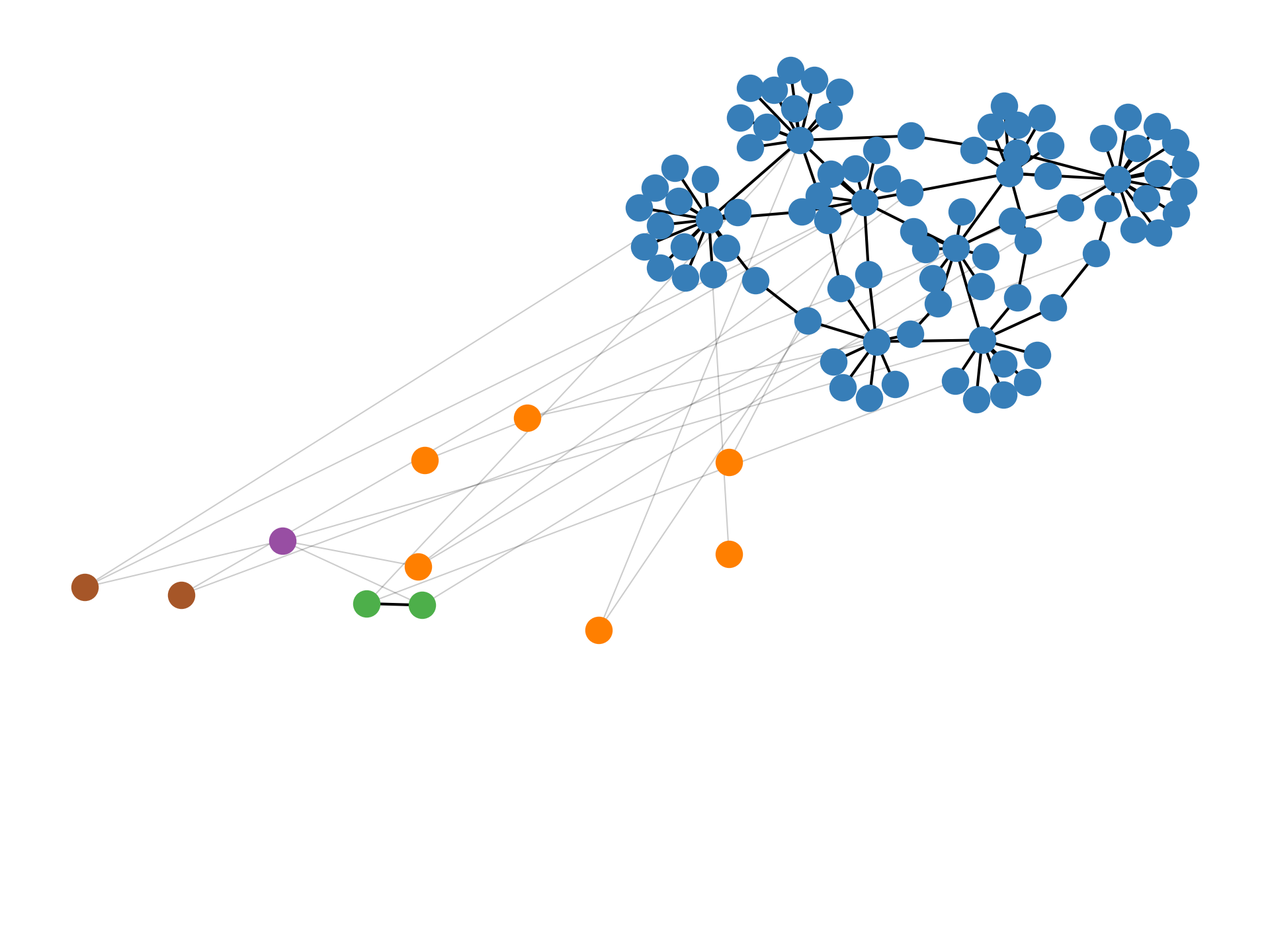}}
%        \caption*{$G_2$}
    \end{subfigure}
    \caption{Word Usage Graph of English \textit{plane} (left), subgraphs for 1st (middle) and 2nd time period (right).}\label{fig:plane}
%\vspace{-15pt}
\end{figure*}

Find an example of an annotated Word Usage Graph from the DWUG EN dataset in Figure \ref{fig:plane}.

\end{document}